\pdfoutput=1

\documentclass{article}
\usepackage{spconf,amsmath,graphicx}
\usepackage{booktabs}
\usepackage{tabularx}

\graphicspath {{Figures/}}

\title{Deep learning-based assessment of tumor-associated stroma for diagnosing breast cancer in histopathology images}
\makeatletter
\def\@name{ \emph{Babak Ehteshami Bejnordi$^{\ast \sharp}$, Jimmy Lin$^{\ddagger}$, Ben Glass$^{\sharp}$, Maeve Mullooly$^{\star}$, Gretchen L Gierach$^{\star}$,}  \\ \emph{{Mark E Sherman}$^{\aleph}$, {Nico Karssemeijer}$^{\ast}$, {Jeroen van der Laak}$^{\ast \dagger}$, {Andrew H Beck}$^{\sharp}$}\\}
\makeatother

\address{$^{\ast}$Diagnostic Image Analysis Group, Radboud University Medical Center, Nijmegen, Netherlands,\\
     $^{\sharp}$Beth Israel Deaconess Medical Center, Harvard Medical School, MA, USA,\\
		 $^{\star}$Division of Cancer Epidemiology and Genetics, National Cancer Institute, NIH, MD, USA,\\
     $^{\dagger}$Dept. of Pathology, Radboud University Medical Center, Nijmegen, Netherlands,\\
     $^{\ddagger}$The Harker School, San Jose, CA, USA,\\
		 $^{\aleph}$Mayo Clinic, Jacksonville, FL, USA
}
\begin{document}
\maketitle
\begin{abstract}
Diagnosis of breast carcinomas has so far been limited to the morphological interpretation of epithelial cells and the assessment of epithelial tissue architecture. Consequently, most of the automated systems have focused on characterizing the epithelial regions of the breast to detect cancer.  In this paper,  we propose a system for classification of hematoxylin and eosin (H\&E) stained breast specimens based on convolutional neural networks that primarily targets the assessment of tumor-associated stroma to diagnose breast cancer patients. We evaluate the performance of our proposed system using a large cohort containing 646 breast tissue biopsies. Our evaluations show that the proposed system achieves an area under ROC of 0.92, demonstrating the discriminative power of previously neglected tumor associated stroma as a diagnostic biomarker.
\end{abstract}
\section{INTRODUCTION}
\label{sec:intro}

Definitive diagnosis and interpretation of breast tissue specimens have so far been based on morphological assessment of epithelial cells. Pathologists traditionally perform a semi-quantitative microscopic assessment of the morphological features of the breast such as tubule formation, epithelial nuclear atypia, and epithelial mitotic activity to detect and characterize breast cancer \cite{PATEY28}. While this assessment has been useful for disease management of cancer patients, the emergence of digital pathology encompassing computerized and computer-aided diagnostics can lead to discovery of valuable prognostic information that provide new insights into the biological factors contributing to breast cancer progression. In \cite{beck2011}, Beck et al. generated new insights into the importance of stromal morphological characteristics as an important prognostic factor in breast cancer that have been previously ignored in the analysis of histopathology images. In addition, several studies have shown that higher stromal density and extent of fibrosis is correlated with increased mammography breast density \cite{sun2013} which confers a 4- to 6-fold risk for breast cancer. These results concord with recent studies which revealed that instead of cancer cells, the surrounding tumor stromal cells known as cancer-associated fibroblasts contribute to cancer progression \cite{shiga2015, kalluri2006}.

However, most of the existing algorithms for breast cancer detection and classification in histology images \cite{dundar2011,Dong2014,Naik2008,Ehteshami16} involve assessment of the morphology and arrangement of epithelial primitives (e.g. nuclei, ducts). Several studies developed automated classification systems based on an initial segmentation of nuclei and extraction of features to describe the morphology of nuclei or their spatial arrangement \cite{dundar2011,Dong2014}. Naik et al. \cite{Naik2008} developed a method for automated detection and segmentation of nuclear and glandular structures for classification of breast cancer histopathology images. While all of the previously mentioned algorithms were designed to classify manually selected regions of interest (mostly selected by expert pathologists), in \cite{Ehteshami16}, we proposed an algorithm for automatic detection of ductal carcinoma in situ (DCIS) that operates at the whole slide level and distinguishes DCIS from a large set of benign disease conditions. 

Unlike the existing work which focused on analysis of epithelial tissue to detect and characterize breast cancer, we sought to develop a novel data-driven system that primarily analyzes stromal morphologic features to discriminate between breast cancer patients and patients with benign breast disease. A crucial step in the development of the existing algorithms has been the design of relevant hand-crafted features. This step is intrinsically intractable for assessing the morphology of tumor stroma in our work. The main reason is that there is no precise definition of the morphological properties of cancer-associated stroma among the pathologists. Moreover, the origin of tumor associated-stromal fibroblasts is not entirely understood \cite{shiga2015}. This motivates the use of machine learning algorithms that can create their own representations for the classification task. Within the field of machine learning, a class of algorithms called deep learning has been very successful in tasks such as image or speech recognition. Deep learning exploits the idea of hierarchical representation learning directly from input data to discover statistical variations in the data. The most successful type of deep learning models for image analysis are convolutional neural networks (CNN). CNNs have also been used to detect cancer areas in breast tissue specimen \cite{cruz201}.

In this paper, we present a new automated system for analyzing H\&E stained breast specimen whole-slide-images (WSI) based on CNNs. Our proposed system distinguishes breast cancer from normal breast tissue based on stromal characteristics of the tissue.
\vspace*{-5 pt}
\section{METHOD}
\label{sec:Method}
\vspace*{-5 pt}
Our proposed model for WSI classification of a breast biopsy specimen consists of several steps. As a pre-processing step, we used a pre-trained network for background/tissue classification. Subsequently, we trained two CNNs. The first one classifies the WSI into epithelium, stroma, and fat. The second operates on the stromal areas resulted from the classification output of the first CNN, to classify the stromal regions as normal stroma or cancer-associated stroma. Two sets of features were extracted from the output of the two CNNs. The first set of features, extracted from the first CNN, characterizes the global tissue amount per class and spatial arrangement of epithelial regions in the WSI. The second set was extracted from the output of the second CNN to characterize several global features related to regions classified as tumor associated stroma. Finally, a random forest classifier was trained using these features to classify the WSI into cancer or non-cancer.
\vspace*{-5 pt}
\subsection{Breast tissue component classification}
Inspired by the success of VGG-net \cite{simonyan2014} which was ranked at the top of ILSVRC 2014 challenge, we trained a VGG-like convolutional neural network with 11 layers. VGG-net uses $3 \times 3$ filters throughout the convolutional layers of the network. Each convolutional layer was followed by a ReLU activation function. We used $2 \times 2$ max-pooling operation after the convolutional layers: 2, 4, 6, and 9. We started with 12 filters, and doubled the number of filters after each max-pooling operation. The densely connected hidden layers have 2048, and 1024 units. To train the network, we replaced the two fully connected layers of our network with $1 \times 1$ convolutions. This is because fully-connected layers require fixed-sized vectors, while convolutional layers accept arbitrary input size; hence, replacing them would allow us to adopt the deep network for images of arbitrary sizes. Hereafter, we denote this model as \textit{CNN I}.
\vspace*{-5 pt}
\subsection{Classification of stromal regions into normal or tumor-associated stroma}
For the second classification task we used the standard 16-layer VGG-net \cite{simonyan2014} which we refer to as \textit{CNN II}. Similar to \textit{CNN I}, we turned this network into an fully convolutional network to allow classification of arbitrary size inputs at test time.

\vspace*{-10 pt}
\subsection{Classification framework and model parameters}

The CNN training procedure for the two networks involves optimizing the multinomial logistic regression objective (softmax) using stochastic gradient descent with Nesterov momentum \cite{Nesterov1983}. The input to both networks is a $224 \times 224$ RGB patch image sampled at the highest magnification. The batch size was set to 128 and 22 for the \textit{CNN I} and \textit{II}, momentum to 0.9. We used L2-regularization ($\lambda _{CNN - I} = 0.003$ and $\lambda _{CNN - II} = 0.0001$) and dropout regularization with ratio 0.5 \cite{srivastava2014} (only applied to the last two layers of the network with $1 \times 1$ convolutions). We used an adaptive learning rate scheme. The learning rate was initially set to 0.01 and then decreased by a factor of 5 if no increase in performance was observed on the evaluation set, over a predefined number of epochs which we refer to as epoch patience ($E_p$). The initial value of $E_p$ was set to 10. We increased this value by 20\% after each drop incidence in the learning rate. This prevented the network from dropping the learning rate too fast at lower learning rates. The weights of our networks were initialized using the strategy by He et al. \cite{he2015}. 

To augment the training set, patches were randomly rotated and flipped. We additionally performed color augmentations, by randomly jittering the hue and saturation of pixels in the HSV color space. To generate the data for each mini-batch, we randomly sampled patches from previously annotated regions for each class with uniform probabilities.

The data for training our \textit{CNN II} exhibits high class imbalance in its distribution (there exists considerably more normal stroma than cancer associated stroma). Although we tried to increase the capacity of the network in learning discriminative features to distinguish the minority class by uniformly sampling the data for each mini-batch, we may fall into the risk of training a very sensitive mode. Because the class distribution in each mini-batch does not represent the actual skewed distribution of the data, a small number of false positives in each mini-batch may translate to vast regions of false positives in the actual WSI. To ameliorate this effect, besides uniform sampling in each mini-batch, we gradually increased the missclassification loss for the normal class. The loss weight factor for the negative samples was initially set to 1, and multiplied by 1.0034 after each epoch (the weight factor becomes 2 by epoch 200). This ensured that the network learns discriminative features from the beginning of the training process and gradually learns the class distribution of the data as well.

\vspace*{-5 pt}
\subsection{Feature extraction and WSI classification}
\textit{CNN I} was applied to the WSI in a sliding window approach to generate a WSI map with three possible labels: epithelium, stroma, and fat. \textit{CNN II} was used to generate WSI likelihood maps representing each pixel's probability of belonging to the tumor associated stroma class. Details of the features extracted from the output of the \textit{CNN I} model are presented in Table 1. These features include: global tissue amount for each class, morphological features of the epithelial areas, and features extracted from Delaunay Triangulation \cite{okabe2009} (built on centroid of epithelial regions) and area-Voronoi diagrams \cite{okabe2009} (generated using the epithelial region areas). We extracted similar features from the thresholded  likelihood maps generated by \textit{CNN II} (T = 0.9) for the connected components labeled as tumor associated stroma.

The resulting feature vector contained 71 features. These features were used to train a random forest classifier with 100 random decision trees. All the parameters including the threshold applied to likelihood maps generated by \textit{CNN II} were tuned using the combination of training and validation sets with cross-validation.

\begin{table}[htbp]
\footnotesize
\centering
\begin{tabularx}{\linewidth}{|l|X|}
\hline
\multicolumn{1}{|c|}{\textit{\textbf{Feature category}}} & \multicolumn{1}{c|}{\textit{\textbf{Feature list}}}                                                   \tabularnewline \hline
Global tissue amount                                               & Total area of epithelium, stroma, and fat and the normalized areas of each tissue class by the total tissue amount.                                                     \tabularnewline \hline
Morphology                                              & $^*$Statistics of the area and eccentricity of epithelial regions                                                                                  \tabularnewline \hline
Delaunay Triangulation                                           &  $^*$Statistics of the number of neighbors for each node and the distances of each node with respect to others                                                                \tabularnewline \hline
Area-Voronoi diagram                                           &  $^*$Statistics of the areas of the Voronoi cells, and the area ratio between the actual epithelial region and its Voronoi zone of influence (ZOI) \cite{bejnordi2013}.                                                               \tabularnewline \hline
\end{tabularx}
\vspace{-1.0em}
\caption{Features extracted from the classification results of \textit{CNN I}. $^*$The statistics we computed are: mean, standard deviation, median, and inter-quartile range.}
\label{table:economicSchools2} 
	\vspace{-0.7em}  
\end{table}
\section{EXPERIMENTS}
\label{sec:format}
\vspace*{-5 pt}

\begin{figure}[htb]
	\centering
		\includegraphics[width=\linewidth]{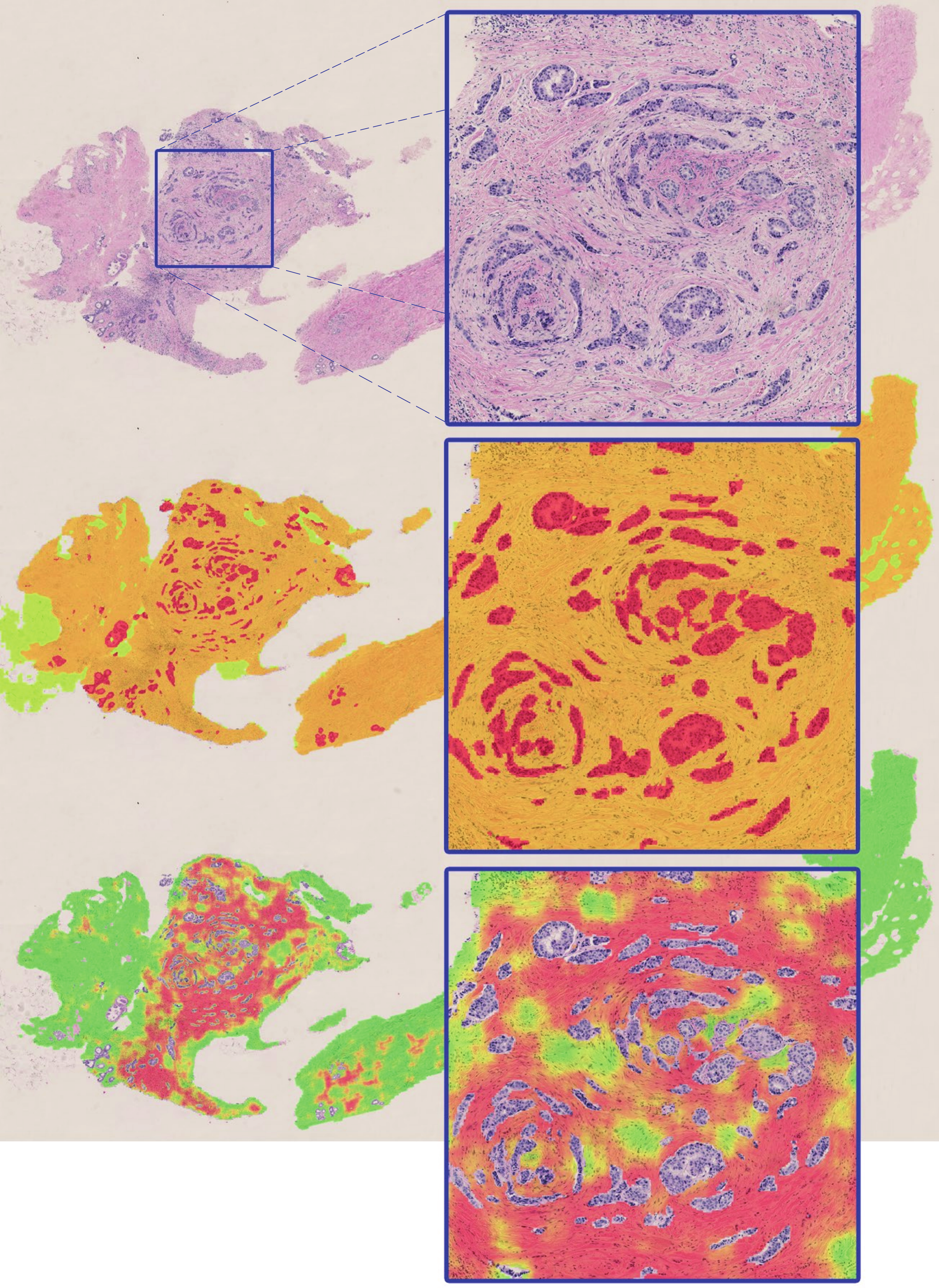}
	\caption{\footnotesize{Sample classification result by \textit{CNN I} (middle image) and \textit{CNN II} (bottom image) for a WSI containing breast cancer. In the middle image, green, orange and red represent fat, stroma, and epithelium, respectively. The bottom image shows the likelihood map representing tumor-stroma probability overlaid on the original image (green represents low probability and red represents high probability of belonging to tumor-associated stroma class).}}
	\label{fig:Sample}
\end{figure}

\subsection{Dataset description}
A total of 646 H\&E stained breast tissue sections obtained from 444 women referred for diagnostic image-guided breast biopsies (including needle core biopsy and vacuum-assisted biopsy) following an abnormal mammogram that were enrolled in the cross-sectional Breast Radiology Evaluation and Study of Tissues (BREAST) Stamp Project \cite{gierach2014} (2007-2010) were included in this analysis. The tissue sections were digitized using Aperio ScanScope CS scanner and Hamamuatsu scanner at 20X magnification, and images have square pixels of size $0.455 \times 0.455 \mu m^2$.

Two trained students annotated a set of epithelial, stromal, and fat regions in 100 WSIs to be used for training and validation of \textit{CNN I}. For the second network we used all the previously annotated stromal regions in normal slides as negative samples. Samples for the tumor associated stroma class were generated under the supervision of a pathologist by annotating stromal regions in the vicinity of epithelial cancer regions.
\subsection{Experimental design}
The dataset was divided into non-overlapping training (270 WSIs with 223 benign disease and 47 invasive cancer), validation (80 WSIs with 65 benign disease and 15 invasive cancer), and testing (296 WSIs with 251 benign disease and 45 invasive cancer) sets. The training and validation sets were used to find the best hyper-parameters for \textit{CNN I} and \textit{II}. The independent test set was used to evaluate the performance of the entire system. For training of both \textit{CNN I} and \textit{II}, we performed two steps of hard negative mining (generating new negative samples from the false positives of the model and retraining).

The final performance of our system was evaluated using receiver-operating characteristic (ROC) analysis on the likelihoods generated by the random forest classifier. Confidence intervals (CI) were generated using patient-stratified bootstrapping with 1000 intervals.
\section{RESULTS AND DISCUSSION}
\vspace*{-6 pt}
\textit{CNN I} achieved a pixel-level accuracy of 95.5\% for classification of tissue into epithelium, stroma, and fat. Fig.~\ref{fig:Sample} shows an example of the result produced by this model on a WSI. \textit{CNN II} achieved a pixel-level binary accuracy of 92.0\% for classifying stroma into normal stroma or tumor-associated stroma. Fig.~\ref{fig:Sample} also shows the results of classification for a slide containing cancer. 

\begin{figure}
	\centering
		\includegraphics[width=\linewidth]{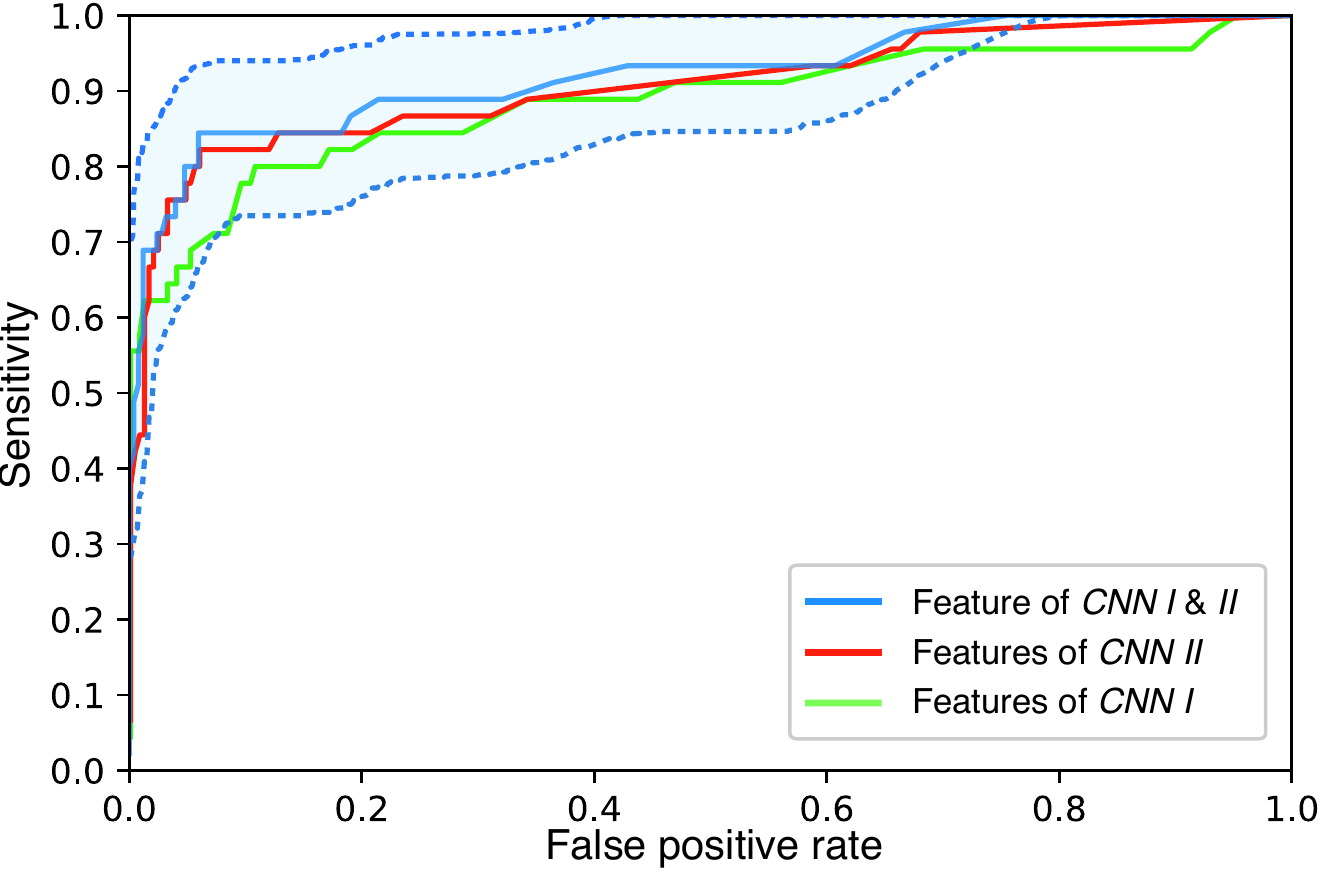}
	\vspace{-1.0em}
	\caption{\footnotesize{The ROC curve of the proposed system. Confidence interval is only shown for the system using both feature sets from \textit{CNN I} and \textit{II}.}}
	\label{fig:ROC_curve}
	\vspace{-1.0em}
\end{figure}

The ROC curve for the final performance of the system is shown in Fig.~\ref{fig:ROC_curve}. The system achieved an AUC of 0.921 (95\% CI 0.862-0.967) at the WSI level for distinguishing cancer from benign breast disease based on combination of both feature sets. The AUC of the system when only considering features from \textit{CNN I} and \textit{CNN II }independently was 0.882 and 0.904, respectively. The results demonstrate that breast cancer can be accurately diagnosed based on the analysis of stromal features alone, suggesting the centrality of alterations to the breast stroma in the process of breast carcinogenesis.

\section{CONCLUSION}
\vspace*{-4 pt}
In this paper we proposed a system for classification of WSIs of breast tissue biopsies. While most previous work has focused on identification of nuclei or glands to characterize abnormal texture patterns in the epithelial cancer regions, we proposed the first system developed based on deep learning techniques to assess the stromal properties of the tissue and investigate the discriminatory power of tumor associated stroma as a diagnostic bio-marker for detecting cancer. These results show that by using deep learning-based techniques, a large amount of information required to discriminate breast cancer from benign breast disease can be obtained from stromal tissue alone. In the future, we will assess the role of tumor associated stroma as a bio-marker to predict breast cancer recurrence and as a predictor of breast cancer development among women diagnosed with benign breast disease.

\section{Acknowledgments}
\vspace*{-4 pt}
This project was funded in part by the Intramural Research Program of the National Cancer Institute, National Institutes of Health, Bethesda, Maryland.

\label{sec:ref}

\bibliographystyle{IEEEbib}
\bibliography{strings,refs}

\end{document}